\documentclass[letterpaper]{article} 
\usepackage[preprint]{aaai2027}  
\usepackage[hyphens]{url}  
\usepackage{graphicx} 
\usepackage{natbib}  
\usepackage{caption} 
\usepackage{algorithm}
\usepackage{algorithmic}

\usepackage{newfloat}
\usepackage{listings}
\DeclareCaptionStyle{ruled}{labelfont=normalfont,labelsep=colon,strut=off} 
\floatstyle{ruled}
\newfloat{listing}{tb}{lst}{}
\floatname{listing}{Listing}

\usepackage{booktabs}
\usepackage{amsmath}
\usepackage{amssymb}

\usepackage{multirow}
\usepackage{colortbl}
\definecolor{selectedrow}{gray}{0.93}

\definecolor{cellworldcomment}{rgb}{0.0, 0.5, 0.5}

\lstdefinestyle{cellworldstyle}{
    language=Python,
    backgroundcolor=\color{white},
    basicstyle=\fontsize{8pt}{9pt}\ttfamily\selectfont,
    commentstyle=\color{cellworldcomment},
    keywordstyle=\bfseries\color{black},
    stringstyle=\color{black},
    showstringspaces=false,
    breaklines=true,
    breakatwhitespace=true,
    numbers=none,
    frame=none,
    columns=fullflexible,
    keepspaces=true,
    mathescape=true,
    xleftmargin=0pt,
    aboveskip=2pt,
    belowskip=2pt
}

\title{CellWorld: From Gene-Level Reconstruction to Latent Cell Prediction in Spatial Transcriptomics Foundation Models}
\author{
    Haiping Liu,
    Qian Zhao,
    Lijing Lin,
    Jingyuan Sun,
    Hongpeng Zhou\corresponding
}
\affiliations{
    University of Manchester\\
    hongpeng.zhou@manchester.ac.uk
}

\begin{document}

\maketitle


\begin{abstract}
This paper shows that latent-space predictive pretraining can provide a scalable route to foundation models for spatial transcriptomics.
Existing spatial transcriptomics foundation models primarily reconstruct masked gene identities or expression values, potentially encouraging the reproduction of assay-specific technical variation and limiting representation transferability.
To avoid directly reconstructing such variation, we shift the prediction target from observed gene measurements to latent cell representations and introduce CellWorld, which predicts the latent representations of masked cells from visible spatial context and a limited partial-expression hint.
We pretrain four CellWorld variants, spanning 5.74M to 94.56M trainable
parameters, on a corpus of 46 million human cells.
Our controlled scaling experiments show that performance improves with model capacity, particularly on spatial tasks, while spatial transfer depends more on sufficient optimization and broad biological source diversity than on cell count alone.
Across four held-out datasets, even CellWorld-Small, with 5.74M trainable
parameters, outperforms every baseline on all 11 linear-probe benchmarks and
all seven fine-tuned spatial benchmarks.
Most notably, a frozen CellWorld-Large pretrained on only 5\% of the corpus with broad biological source coverage outperforms every fully fine-tuned baseline across all seven spatial benchmarks.
Code is available at \url{https://github.com/UoM-HealthAI/CellWorld}.
\end{abstract}


\section{Introduction}
Spatial transcriptomics (ST) measures gene expression while preserving tissue spatial organization, enabling cells to be characterized within their local tissue context~\citep{staahl2016visualization,merfish}. 
The rapidly growing scale and diversity of ST datasets create an opportunity to learn transferable representations across datasets and downstream tasks through large-scale pretraining.

The first generation of ST foundation models~\citep{nicheformer, cellplm, brainbeacon, heist, scgpt-spatial} has pursued this opportunity primarily through masked reconstruction of gene-level observations. 
However, ST measurements are sparse and heterogeneous across platforms owing to differences in targeted gene panels, measurement noise, and batch effects. Direct reconstruction may therefore encourage models to reproduce assay-specific variation alongside biological signal.

Latent prediction instead follows the principle that
\emph{transferable representations can be learned by predicting the abstract states of unobserved inputs rather than reconstructing raw observations}~\citep{data2vec}.
Joint-embedding predictive architectures (JEPAs) have demonstrated the potential of this principle in vision, where models predict the latent representations of masked target regions from visible context~\citep{i-jepa,v-jepa}.
However, directly applying this formulation to ST is nontrivial.
Image regions typically exhibit strong local continuity with their surroundings, whereas ST slides comprise discrete cells whose identities and expression states can differ even across otherwise similar spatial neighborhoods.
Visible spatial context alone may therefore not uniquely determine the identity and molecular state of a masked cell.
We refer to this underdetermination as cell-level target \emph{ambiguity}.
To date, it remains unclear whether and under what formulation latent prediction can serve as a reliable and scalable pretraining objective for ST foundation models.

To address this gap, we introduce CellWorld, a large-scale ST foundation model pretrained through latent cell prediction.
Given a local patch of spatially neighboring cells, CellWorld maps each cell's gene-expression profile and metadata to a cell token and models cell--cell interactions using spatial Transformers.
It randomly designates a subset of cells as masked targets and uses the remaining cells as visible context.
A context encoder processes only the visible cell tokens, whereas an EMA-updated target encoder processes the complete patch to generate a latent target for each masked cell.
A spatial predictor is trained to recover each latent target from the encoded visible context and the corresponding target coordinate.
To resolve cell-level target ambiguity, the predictor is additionally conditioned on a limited partial-expression hint from the masked cell.
After pretraining, the target encoder and predictor are discarded, while the cell tokenizer and context encoder are retained for downstream tasks.

\begin{figure*}[t]
    \centering
    \includegraphics[width=0.8\textwidth]{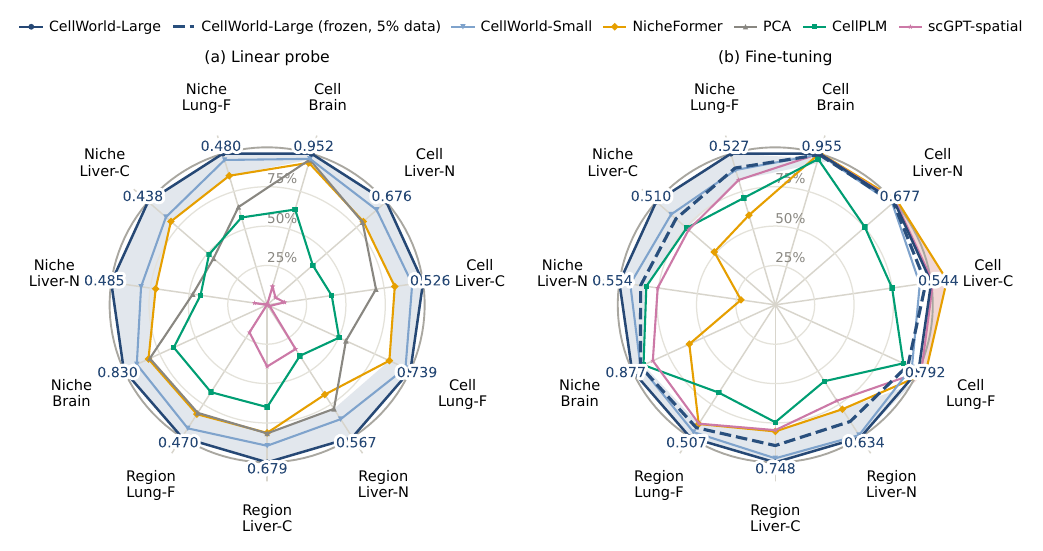}
    \caption{
    Comparison with existing methods across 11 held-out task--dataset pairs under
    (a) linear probing and (b) fine-tuning.
    Each axis is normalized to the full-data CellWorld-Large score under the
    corresponding protocol (100\%); numeric annotations give its absolute scores.
    The dashed contour in (b) shows the frozen CellWorld-Large pretrained on a
    broadly sampled 5\% corpus subset, normalized to the fine-tuned full-data model.
    Higher is better.
    }
    \label{fig:baseline_radar}
\end{figure*}

We pretrain four CellWorld variants, namely Small, Base, Large, and Huge, on 46 million human cells spanning three platforms and 11 organs.
As shown in Figure~\ref{fig:baseline_radar}, even CellWorld-Small, with only
5.74M trainable parameters, outperforms every existing method on all 11
benchmarks under linear probing and all seven spatial benchmarks under
fine-tuning, with larger variants further extending this lead.
Our controlled scaling experiments show continued gains with model capacity, particularly on spatial tasks.
Data scaling shows that transferable information about cell identity can be learned from relatively little pretraining data, whereas spatial transfer is governed less by raw cell count than by \emph{broad biological source diversity and sufficient optimization}.
Most notably, a frozen CellWorld-Large pretrained on only 5\% of the corpus with broad biological source coverage outperforms every fully fine-tuned baseline on all seven spatial benchmarks.

\section{Related Work}
\paragraph{ST foundation models.}
Nicheformer~\citep{nicheformer} predicts masked gene identities from expression-ranked gene sequences, whereas scGPT-spatial~\citep{scgpt-spatial}, CellPLM~\citep{cellplm}, and BrainBeacon~\citep{brainbeacon} reconstruct masked expression values.
HEIST~\citep{heist} additionally incorporates spatial and contrastive objectives.
SToFM~\citep{stofm} reconstructs masked, expression-derived cell embeddings produced by a domain-adapted single-cell encoder, alongside a spatial reconstruction objective, and thus remains reconstruction-based despite operating in an embedding space.
In contrast, CellWorld is pretrained to predict contextualized latent representations of masked cells.

\paragraph{Latent predictive representation learning.}
Latent prediction has emerged as an alternative to reconstructive self-supervision across visual modalities. 
BYOL~\citep{byol} trains an online network to predict the representation produced by an exponential-moving-average target network under a different augmentation. 
I-JEPA~\citep{i-jepa} predicts latent representations of masked image regions from visible context, while subsequent work has explored and extended this paradigm across diverse architectures, modalities, and application domains~\citep{v-jepa,v-jepa2.1,vl-jepa,point-jepa,lejepa,lejepa-world,genejepa,celljepa}.
Concurrent work, ST-JEPA~\citep{st-jepa}, applies latent prediction to masked gene tokens in sequences constructed from local cellular neighborhoods, focusing on representation learning evaluated through clustering rather than large-scale cross-dataset pretraining and transfer.
CellWorld instead represents each cell as a single token, models cell--cell interactions with a spatial encoder, and pretrains at scale by predicting contextualized latent representations of masked cells for transfer across diverse datasets and downstream tasks.

\section{Method}
\subsection{Spatial Sampling Strategy}
\label{subsec:sampling-strategy}
We construct 4{,}096-cell local patches through a three-stage spatial sampling procedure.
First, we partition each slide into spatially coherent subslides using \(k\)-means on the two-dimensional cell coordinates, with \(K_{\mathrm{sub}}=\max(1,\lfloor n/50{,}000 \rfloor)\) subslides for a slide containing \(n\) cells.
Within each subslide \(s\), farthest-point sampling (FPS) selects \(K_s=2\lceil n_s/4096 \rceil\) patch centres, where \(n_s\) is the number of cells in the subslide.
FPS is rerun at each epoch, and the 4{,}096 cells nearest to each centre form a training sample.

\subsection{CellWorld Pretraining Framework}
\begin{figure}[ht]
    \centering
    \includegraphics[width=0.8\columnwidth]{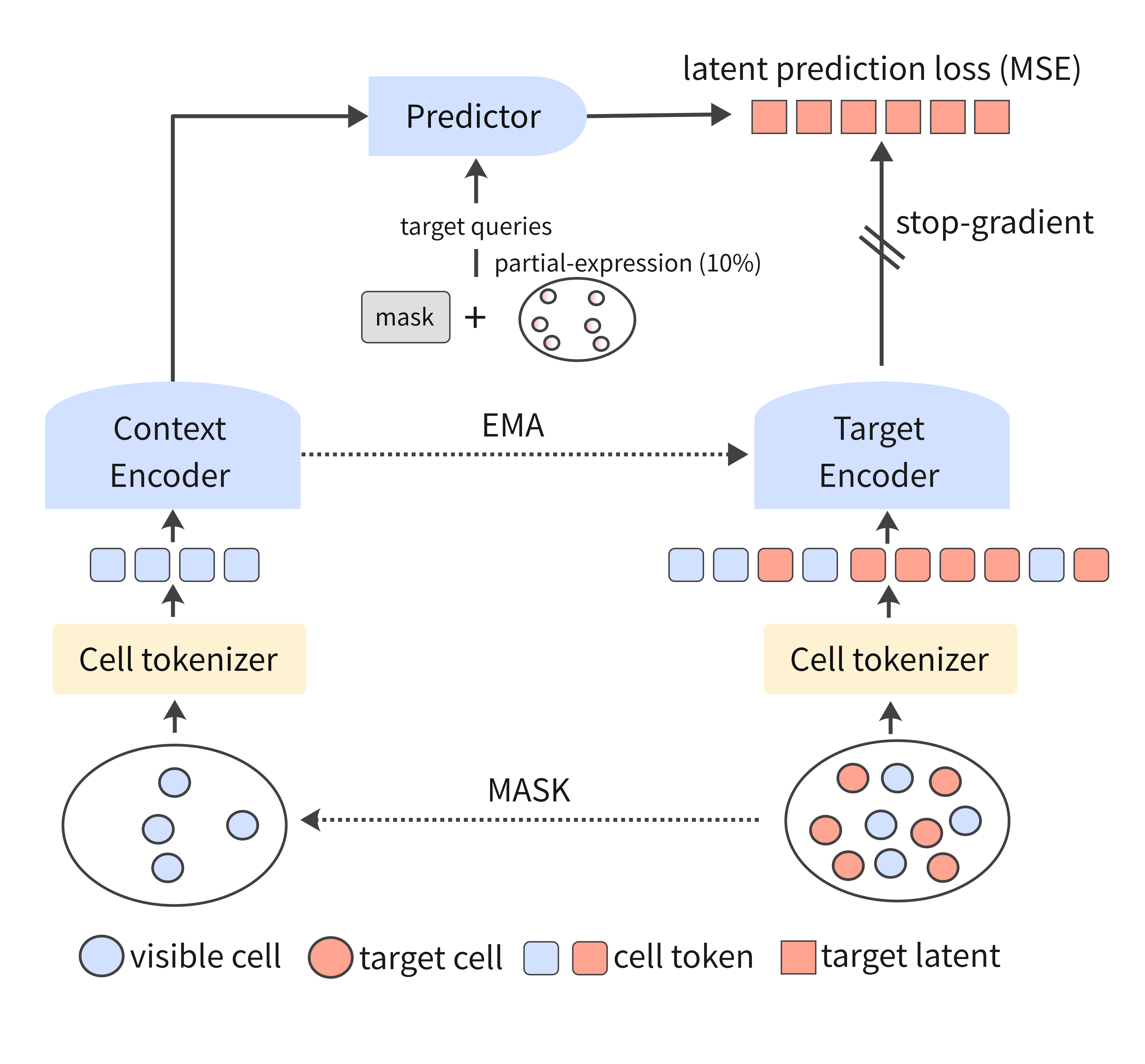}
    \caption{Overview of the CellWorld pretraining architecture.}
    \label{fig:cellworld-overview}
\end{figure}

Figure~\ref{fig:cellworld-overview} illustrates the overall architecture.
Within each sampled spatial patch, CellWorld converts gene-expression profiles into cell tokens, models their interactions using spatial attention, and predicts the latent representations of masked target cells from the encoded visible context and limited partial-expression hints.
We detail each component below.

\paragraph{Cell tokenization.}
We first map each cell's gene symbols to a unified vocabulary of 19{,}227 genes constructed from the pretraining corpus, with a single embedding for each gene shared across platforms.
For each cell \(i\), \(\mathcal{G}_i\) contains up to \(K=512\) genes with the highest expression values; this retains the complete assayed panel for 95\% of cells.
CellWorld then constructs the cell token by aggregating their embeddings according to their expression values and adding organ and platform embeddings:
\begin{equation}
    \mathbf{u}_i
    =
    \sum_{g\in\mathcal{G}_i} x_{ig}\mathbf{e}_g
    +
    \mathbf{e}^{\mathrm{organ}}_i
    +
    \mathbf{e}^{\mathrm{platform}}_i.
\end{equation}
Here, \(x_{ig}\) is the expression value of gene \(g\),
\(\mathbf{e}_g\in\mathbb{R}^{d}\) is its gene embedding, and
\(\mathbf{e}^{\mathrm{organ}}_i\) and
\(\mathbf{e}^{\mathrm{platform}}_i\) are metadata embeddings.
The context encoder, target encoder, and hint-conditioned spatial predictor share the same cell tokenizer.

\paragraph{Spatial attention.}
Because coordinate systems and spatial scales vary across slides, we avoid absolute positional embeddings and instead encode relative spatial relationships through attention biases.
Specifically, before masking, coordinates are normalized within each spatial patch as
\(\tilde{\mathbf{s}}_i=(\mathbf{s}_i-\bar{\mathbf{s}})/\delta\),
where \(\bar{\mathbf{s}}\) is the patch center and
\(\delta=\max_j\|\mathbf{s}_j-\bar{\mathbf{s}}\|_\infty\)
is the largest absolute coordinate displacement.
CellWorld then adopts a two-dimensional variant of ALiBi~\cite{alibi}, which converts pairwise distances between normalized coordinates into additive self-attention biases using fixed, head-specific slopes.
For attention head \(h\), the bias is
\(B_{ij}^{(h)}=-m_h\|\tilde{\mathbf{s}}_i-\tilde{\mathbf{s}}_j\|_2\)
and is added to the standard attention logits:
\begin{equation}
    \operatorname{Attn}^{(h)}
    =
    \operatorname{softmax}\!\left(
        \frac{\mathbf{Q}^{(h)}\mathbf{K}^{(h)\top}}{\sqrt{d_h}}
        + \mathbf{B}^{(h)}
    \right)\mathbf{V}^{(h)}.
\end{equation}
Here, \(m_h\) controls the effective spatial scale: larger values emphasize nearby cells, whereas smaller values preserve broader spatial context.
This mechanism is used by the context encoder, target encoder, and predictor.

\paragraph{Random masking and context encoder.}
For each spatial patch, we apply random masking by uniformly sampling without replacement a fraction \(r=0.6\) of cells as prediction targets \(\mathcal{M}\), with the remaining cells forming the visible context \(\mathcal{C}\).
The target set is resampled each time a patch is presented during pretraining.
Target cells are removed before context encoding rather than replaced with learnable mask tokens.
The context encoder \(f_\theta\) therefore processes only the visible cell tokens and their normalized coordinates:
\begin{equation}
    \mathbf{z}_{\mathcal{C}}
    =
    f_\theta\left(
        \mathbf{u}_{\mathcal{C}},
        \tilde{\mathbf{s}}_{\mathcal{C}}
    \right).
\end{equation}

\paragraph{Target encoder.}
The target encoder \(f_{\bar{\theta}}\) mirrors the context encoder but processes the complete patch, with its outputs at the target positions defining the latent prediction targets:
\begin{equation}
    \mathbf{z}_{\mathcal{M}}
    =
    \operatorname{sg}
    \left[
        f_{\bar{\theta}}
        \left(
            \mathbf{u}_{\mathcal{C}\cup\mathcal{M}},
            \tilde{\mathbf{s}}_{\mathcal{C}\cup\mathcal{M}}
        \right)_{\mathcal{M}}
    \right],
\end{equation}
where \(\operatorname{sg}\) denotes stop-gradient.
These representations serve as latent targets that the predictor is trained to recover.

\paragraph{Hint-conditioned spatial predictor.}
To reduce cell-level target ambiguity, we condition each target query on a limited partial-expression hint.
For each target cell \(i\), we define its retained non-zero gene set as
\(\mathcal{G}_i^+=\{g\in\mathcal{G}_i:x_{ig}>0\}\)
and independently include each gene with probability \(\kappa=0.10\), using
\(\xi_{ig}\sim\operatorname{Bernoulli}(\kappa)\).
The resulting hint representation is
\begin{equation}
    \mathbf{h}_i
    =
    \sum_{g\in\mathcal{G}_i^+}
    \xi_{ig}x_{ig}\mathbf{e}_g
    +
    \mathbf{e}^{\mathrm{organ}}_i
    +
    \mathbf{e}^{\mathrm{platform}}_i.
\end{equation}

The hint is used exclusively by the predictor and is not passed to either encoder.
Each target query is
\(\mathbf{q}_i=\mathbf{m}+W_{\mathrm{cp}}\mathbf{h}_i\),
where \(\mathbf{m}\) is a learnable mask token and
\(W_{\mathrm{cp}}\) projects the hint into the predictor space.
The same projection maps the visible context representations into this space, after which they are concatenated with the target queries.
Using the normalized coordinates of both visible and target cells, the spatial predictor \(p_\phi\) produces
\begin{equation}
    \hat{\mathbf{z}}_{\mathcal{M}}
    =
    W_{\mathrm{out}}\,
    p_\phi
    \left(
        [W_{\mathrm{cp}}\mathbf{z}_{\mathcal{C}};\mathbf{q}_{\mathcal{M}}],
        [\tilde{\mathbf{s}}_{\mathcal{C}};\tilde{\mathbf{s}}_{\mathcal{M}}]
    \right)_{\mathcal{M}},
\end{equation}
where \(W_{\mathrm{out}}\) maps the predictor outputs back to the target-representation dimension.
The predicted representations \(\hat{\mathbf{z}}_{\mathcal{M}}\) are matched against the target representations \(\mathbf{z}_{\mathcal{M}}\) under the training objective described next.

\paragraph{Training objective.}
We minimize the mean squared error between the predicted and target representations over all target cells:
\begin{equation}
    \mathcal{L}_{\mathrm{pred}}
    =
    \frac{1}{|\mathcal{M}|d}
    \sum_{i\in\mathcal{M}}
    \left\|
        \hat{\mathbf{z}}_i-\mathbf{z}_i
    \right\|_2^2,
\end{equation}
where \(d\) is the target representation dimension.
The target encoder is initialized from the context encoder and updated only through an exponential moving average:
\(
    \bar{\theta}
    \leftarrow
    \tau\bar{\theta}
    +
    (1-\tau)\theta,
\)
where \(\tau\) is the momentum coefficient.
All other components, including the cell tokenizer, context encoder, predictor, and projection layers, are optimized through backpropagation.

\subsection{Model Configurations and Pretraining}
\paragraph{Model configurations.} We instantiate CellWorld at four
scales---Small, Base, Large, and Huge---by varying the width, depth, and number
of attention heads in the context encoder and spatial predictor. The resulting
models span 5.74M to 94.56M trainable parameters. Detailed architectural
configurations are provided in Table~\ref{tab:model-configs}.

\begin{table}[t]
    \centering
    \footnotesize
    \setlength{\tabcolsep}{3pt}
    \renewcommand{\arraystretch}{1.08}
    \begin{tabular}{@{}lcccc@{}}
        \toprule
        Model
        & Enc. (W/L/H)
        & Pred. (W/L/H)
        & Train. (M)
        & Total (M) \\
        \midrule
        Small & 192/6/6   & 96/3/3  & 5.74  & 7.52 \\
        Base  & 384/8/6   & 192/3/3 & 17.90 & 27.37 \\
        Large & 512/12/8  & 256/3/4 & 36.93 & 62.17 \\
        Huge  & 768/16/12 & 384/3/6 & 94.56 & 170.19 \\
        \bottomrule
    \end{tabular}
    \caption{Architectural configurations of the CellWorld model family. Enc. and Pred. denote the context encoder and spatial predictor, respectively; W/L/H denotes embedding width, number of layers, and number of attention heads. Trainable parameter counts exclude the EMA target encoder, whereas total parameter counts include it.}
    \label{tab:model-configs}
\end{table}

\paragraph{Pretraining recipe and compute.}
We pretrain CellWorld on the full corpus for 10{,}000 optimization steps with a global batch size of 256 using BF16 precision and AdamW.
The learning rate is linearly warmed up to \(3\times10^{-3}\) over the first 1{,}000 steps and then decayed using a cosine schedule.
The EMA momentum of the target encoder increases linearly from 0.9990 to 0.9999 throughout pretraining.
We use random masking with a mask ratio of 0.6 and a target-cell hint ratio of 0.10.
All four model scales are trained on AMD MI250X accelerators, with compute allocation scaled from 8 to 32 GPU compute dies (GCDs).
The same recipe remains numerically stable across all model scales; additional optimization and compute details are provided in Appendix~C.

\section{Experiments}
\subsection{Experimental Setup}
To pretrain CellWorld, we assemble a corpus of 46 million human cells spanning 11 organs and three high-resolution spatial transcriptomics platforms---MERFISH, Xenium, and CosMx. We additionally curate four datasets exclusively for downstream evaluation: MERFISH human brain, CosMx normal liver, CosMx liver cancer, and Xenium lung fibrosis. All four datasets are excluded from pretraining for both CellWorld and the baseline models. 
Expression counts in both the pretraining corpus and downstream datasets are normalized to 10{,}000 counts per cell and log-transformed as \(\log(1+x)\). 
Each downstream dataset is additionally restricted to 300 highly variable genes selected using all of its cells; no other preprocessing is applied.
Detailed dataset statistics are provided in Appendix~A.

We consider three downstream tasks: cell annotation, which evaluates cell-level biological identity; region prediction, which assigns cells to annotated tissue regions; and niche-composition prediction, which estimates the local proportions of neighboring cell types. We use macro-F1 for cell annotation and region prediction and the Pearson correlation coefficient (PCC) for niche-composition prediction. Throughout the tables, these tasks are denoted Cell, Region, and Niche, respectively.

Our experiments investigate CellWorld's main design choices under controlled settings, scaling behavior with model capacity and pretraining data scale, and performance relative to existing methods. Each task is evaluated on all held-out datasets for which the required annotations are available. The controlled analyses use linear probing only, whereas the scaling and comparison experiments use both linear probing and full fine-tuning. 
Under linear probing, all pretrained encoders are frozen, and the same task-specific head architecture and evaluation protocol are used for every CellWorld variant and external baseline.
CellWorld fine-tuning uses a single shared configuration across all model- and data-scaling conditions, whereas external baselines use architecture-specific best-effort configurations. 
Complete configurations are provided in Appendices~B--D.

We split each downstream dataset by FOV rather than by individual cells, ensuring that cells from the same FOV do not appear in both training and validation sets, thereby reducing spatial information leakage. 
For the controlled-design and scaling experiments, we average each dataset's best validation score across three downstream seeds and then compute unweighted means across applicable datasets.
Comparisons with existing methods instead report every task--dataset pair, averaged over the three downstream seeds.
Details of the FOV-level train--validation splits, task construction, and metric computation are provided in Appendix~B, with complete dataset-level results for the controlled-design and scaling experiments in Appendix~E.

\subsection{Controlled Design Analysis}
We conduct controlled comparisons using CellWorld-Base pretrained on the full corpus with the default recipe, varying one design factor at a time while holding all other settings fixed.
For the target-cell-hint analysis, we additionally include matched CellWorld-Large runs to assess whether the effect depends on model capacity.
Additional ablations of model components, including metadata embeddings and 2D-ALiBi, are provided in Appendix~E.

\paragraph{Masking strategy.}
Masking controls both the spatial organization of prediction targets and the visible context. We compare uniform random masking with spatial block masking, which selects four connected target regions on the spatial \(k\)-nearest-neighbour graph, following vision JEPA models~\cite{i-jepa,v-jepa}. To separate masking strategy from mask ratio, we evaluate both at \(r \in \{0.4, 0.6, 0.8\}\), where \(r\) is the fraction of cells withheld from the context encoder as prediction targets. As shown in Table~\ref{tab:masking-strategy}, random masking outperforms block masking in eight of nine task-level comparisons; the only exception is region prediction at \(r=0.6\), with a difference below 0.001 before rounding. Random masking also performs better in 27 of 33 matched dataset-level comparisons (Appendix~E). We therefore adopt random masking with \(r=0.6\), as performance varies only marginally across random-mask ratios.

\begin{figure}[t]
    \centering
    \includegraphics[width=1\columnwidth]{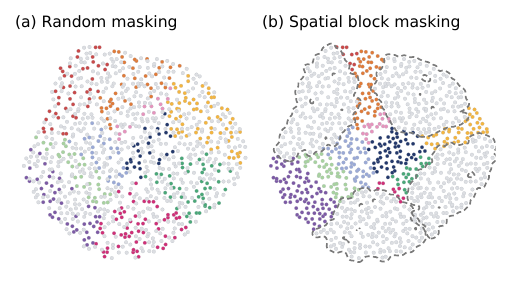}
    \caption{
    Illustration of (a) random masking and (b) spatial block masking at the same 60\% mask ratio. Colored cells represent visible context cells, with colors indicating different cell populations; gray cells denote masked targets, and dashed contours mark the sampled spatial blocks.
    }
    \label{fig:masking-schemes}
\end{figure}

\begin{table}[t]
    \centering
    \footnotesize
    \setlength{\tabcolsep}{3.5pt}
    \renewcommand{\arraystretch}{1.08}
    \begin{tabular}{@{}c|cc|cc|cc@{}}
        \toprule
        & \multicolumn{2}{c}{Cell}
        & \multicolumn{2}{c}{Region}
        & \multicolumn{2}{c}{Niche} \\
        \cmidrule(lr){2-3}
        \cmidrule(lr){4-5}
        \cmidrule(lr){6-7}
        Mask ratio
        & Random & Block
        & Random & Block
        & Random & Block \\
        \midrule
        0.4
        & 0.719 & 0.716
        & 0.544 & 0.543
        & 0.533 & 0.532 \\
        0.6
        & 0.719 & 0.717
        & 0.539 & 0.540
        & 0.531 & 0.526 \\
        0.8
        & 0.719 & 0.715
        & 0.537 & 0.533
        & 0.531 & 0.525 \\
        \bottomrule
    \end{tabular}
    \caption{Comparison of random and spatial block masking.}
    \label{tab:masking-strategy}
\end{table}

\paragraph{Target-cell hint.} 
We next examine how partial gene-expression hints from target cells affect latent prediction.
As shown in Table~\ref{tab:hint}, CellWorld-Base without hints performs best on both spatial tasks, whereas a hint ratio of 0.10 performs best on cell annotation and increasing it to 0.50 reduces performance across all three tasks. 
Predictor diagnostics in Appendix~E nevertheless show that the no-hint predictor produces nearly identical outputs for different targets within the same spatial patch, approximating a patch-conditioned mean, while a 0.10 hint restores target-specific variation. 
This reveals a trade-off between distinguishing individual targets and preserving reliance on spatial context. Because downstream evaluation uses the context encoder, the degeneration remains confined to the predictor at the Base scale and does not impair spatial transfer.

\begin{figure}[t]
    \centering
    \includegraphics[width=\columnwidth]
    {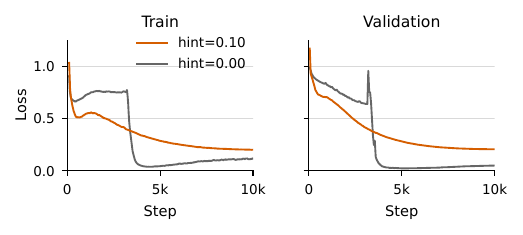}
    \caption{Training and validation loss trajectories for CellWorld-Large with hint ratios of 0 and 0.10.}
    \label{fig:hint-training-dynamics}
\end{figure}

At the Large scale, however, the same predictor degeneration is no longer benign. Figure~\ref{fig:hint-training-dynamics} shows that no-hint training collapses after approximately 3,000 steps, and Table~\ref{tab:hint} shows substantial degradation across all three tasks. 
In contrast, the 0.10-hint model avoids this predictor degeneration, remains stable throughout training, and outperforms its no-hint counterpart on every task.
We therefore adopt a hint ratio of 0.10 to stabilize scaling while limiting target information.

\begin{table}[t]
    \centering
    \footnotesize
    \renewcommand{\arraystretch}{1.08}
    \begin{tabular*}{\columnwidth}{@{\extracolsep{\fill}}lc|c|c|c@{}}
        \toprule
        Scale
        & Hint ratio
        & Cell
        & Region
        & Niche \\
        \midrule
        Base
        & 0.00
        & 0.714
        & \textbf{0.571}
        & \textbf{0.551} \\
        Base
        & 0.10
        & \textbf{0.719}
        & 0.539
        & 0.531 \\
        Base
        & 0.50
        & 0.703
        & 0.530
        & 0.522 \\
        \midrule
        Large
        & 0.00
        & 0.705
        & 0.502
        & 0.462 \\
        Large
        & 0.10
        & \textbf{0.723}
        & \textbf{0.572}
        & \textbf{0.558} \\
        \bottomrule
    \end{tabular*}
    \caption{Downstream performance across target-cell hint ratios at Base and Large model scales. Bold indicates the best result within each scale.}
    \label{tab:hint}
\end{table}

\paragraph{Prediction objective.}
We next compare latent target prediction with a matched MAE-style control~\cite{mae}, in which the predictor serves as a decoder that reconstructs selected expression values of masked cells. Both objectives use a mask ratio of 0.6 and a target-cell hint ratio of 0.10, with all other settings matched; we tune the peak learning rate for expression reconstruction separately to ensure stable training (Appendix~D). 
As shown in Table~\ref{tab:model-properties}, expression reconstruction achieves comparable cell annotation performance but performs worse on both spatial tasks. This advantage holds for latent prediction in six of the seven spatial task--dataset pairs (Appendix~E), supporting it as a more effective objective for learning spatially contextual representations.

\paragraph{Spatial context.}
We test whether CellWorld uses spatial neighborhood structure by permuting cell expression profiles across fixed coordinate slots within each subslide during both pretraining and downstream inference, while preserving the coordinate-defined KNN graph. As shown in Table~\ref{tab:model-properties}, this permutation leaves cell annotation essentially unchanged but reduces performance on both spatial tasks, indicating that CellWorld exploits the correspondence between cellular states and their spatial neighborhoods.

\begin{table}[t]
    \centering
    \footnotesize
    \renewcommand{\arraystretch}{1.08}
    \begin{tabular*}{\columnwidth}{@{\extracolsep{\fill}}l|c|c|c@{}}
        \toprule
        Configuration
        & Cell
        & Region
        & Niche \\
        \midrule
        Full CellWorld
        & 0.719
        & \textbf{0.539}
        & \textbf{0.531} \\
        Expression reconstruction
        & \textbf{0.721}
        & 0.535
        & 0.520 \\
        Permuted spatial assignment
        & 0.718
        & 0.529
        & 0.511 \\
        \bottomrule
    \end{tabular*}
    \caption{Controlled comparisons of prediction objective and spatial context.
    Bold indicates the best result in each column.}
    \label{tab:model-properties}
\end{table}

\subsection{Scaling Behavior}

\begin{figure}[t]
    \centering
    \includegraphics[width=0.9\columnwidth]{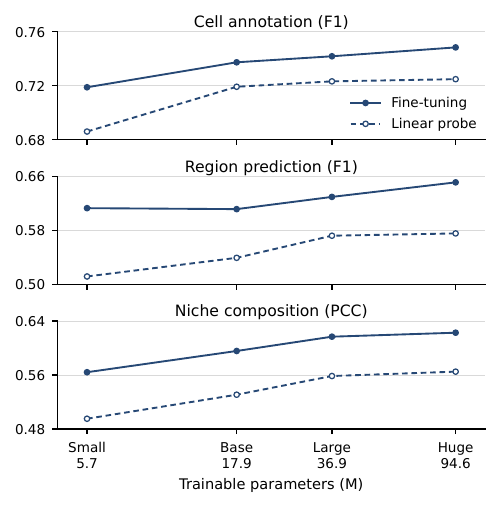}
    \caption{CellWorld model scaling under linear probing and fine-tuning
    versus trainable parameter count (log scale).}
    \label{fig:model-scaling}
\end{figure}

\paragraph{Model capacity.}
We first scale CellWorld from Small to Huge while keeping the corpus and recipe fixed.
Overall, as shown in Figure~\ref{fig:model-scaling}, CellWorld exhibits clear scaling behavior under both evaluation protocols, with CellWorld-Huge achieving the highest task-level performance across all three tasks.
Under linear probing, performance improves monotonically across all three tasks, indicating that larger models learn more transferable pretrained representations.
Under fine-tuning, cell annotation and niche-composition prediction also improve monotonically, while region prediction is the only exception, decreasing slightly from Small to Base before improving at Large and Huge.

Across tasks, the gains in cell annotation are concentrated in scaling from Small to Base, with only modest improvements thereafter. This early saturation suggests that Base already provides sufficient capacity to capture most cell-type-discriminative information. In contrast, region prediction and niche-composition prediction show substantial gains from Base to Large. Beyond Large, gains generally taper, while region prediction under fine-tuning continues to improve substantially. 

\paragraph{Data scaling.}
To characterize how CellWorld scales with pretraining data, we pretrain CellWorld-Large on 5\%, 25\%, and 100\% of the corpus under the default recipe and 10{,}000-step schedule.
The smaller subsets are constructed by distributed subsampling of local patches across source slides, reducing the total cell count while largely preserving biological source diversity across organs and slides.
Both distributed subsets retain all three platforms and 11 organs; the 25\% subset includes all 72 slides, while the 5\% subset retains 64.
Remarkably, Table~\ref{tab:data-scaling-budgets} shows that the 25\% subset matches or slightly outperforms the full corpus across tasks, while the 5\% subset incurs only marginal overall degradation.
At face value, these results suggest that transfer performance saturates with only a small fraction of the pretraining cells.

However, this apparent saturation may reflect two factors: under the fixed 10{,}000-step schedule, smaller subsets revisit the same cells and local contexts more frequently, while distributed subsampling preserves broad biological source diversity.
We first test repeated exposure by keeping the sampling strategy fixed and scaling the pretraining steps in proportion to data size: 500, 2{,}500, and 10{,}000 steps for the 5\%, 25\%, and 100\% subsets, respectively.
This approximately equalizes the number of visits per local context across data scales.
Surprisingly, compared with the same subsets trained for 10{,}000 steps, the proportionally scaled models retain comparable cell annotation performance but show marked declines on both spatial tasks under linear probing and fine-tuning.
These results suggest that cell annotation saturates after relatively few steps, whereas spatial transfer improves markedly with additional optimization even on the same cells and local contexts.

We next test the second factor by replacing distributed patch subsampling with whole-slide sampling while retaining the 10{,}000-step schedule, thereby concentrating each data fraction in fewer slides and organs.
At 25\% data, reducing coverage from 72 slides and 11 organs to 11 slides and six organs leaves performance largely unchanged.
Further reducing coverage to four slides and two organs preserves cell annotation but consistently lowers Region and Niche performance under both evaluation protocols.
Notably, despite containing only one-fifth as many cells, the broadly sampled 5\% subset matches or slightly outperforms the concentrated 25\% subset on both spatial tasks under both protocols.
Most strikingly, at a matched 5\% data budget, concentrating the data from 64 slides and 11 organs to four slides and three organs substantially reduces all three linear-probe scores.
Fine-tuning nearly recovers cell annotation, but substantial gaps remain on both spatial tasks.
Thus, biological source diversity becomes particularly important for spatial transfer in the low-data regime.

Taken together, these experiments show that cell count alone does not determine the effective pretraining scale.
Optimization and biological source diversity play complementary roles: repeated exposure allows CellWorld to extract more transferable spatial structure from the available cells, whereas source diversity determines the breadth of spatial variation available to learn.
Under a limited cell budget, distributing cells across organs and slides can therefore be more valuable than adding cells from a small number of sources, particularly for spatial transfer.

\begin{table*}[t!]
    \centering
    \begin{minipage}{0.86\textwidth}
        \centering
        \footnotesize
        \setlength{\tabcolsep}{2.5pt}
        \setlength{\arrayrulewidth}{0.35pt}
        \renewcommand{\arraystretch}{1.04}

        \begin{tabular*}{\linewidth}{
            @{\extracolsep{\fill}}cccc|ccc|ccc@{}
        }
            \toprule
            Data (\%)
            & Slides
            & Organs
            & Steps (k)
            & Cell (LP)
            & Region (LP)
            & Niche (LP)
            & Cell (FT)
            & Region (FT)
            & Niche (FT)
            \\
            \midrule

            100 & 72 & 11 & 10
            & 0.723 & 0.572 & 0.558
            & 0.742 & 0.630 & 0.617
            \\
            25 & 72 & 11 & 10
            & 0.724 & 0.569 & 0.559
            & 0.743 & 0.637 & 0.617
            \\
            5 & 64 & 11 & 10
            & 0.720 & 0.565 & 0.551
            & 0.740 & 0.631 & 0.610
            \\

            \midrule

            25 & 72 & 11 & 2.5
            & 0.725 & 0.542 & 0.532
            & 0.737 & 0.586 & 0.578
            \\
            5 & 64 & 11 & 0.5
            & 0.720 & 0.531 & 0.514
            & 0.741 & 0.589 & 0.568
            \\

            \midrule

            25 & 11 & 6 & 10
            & 0.724 & 0.568 & 0.562
            & 0.743 & 0.637 & 0.610
            \\
            25 & 4 & 2 & 10
            & 0.724 & 0.563 & 0.549
            & 0.741 & 0.628 & 0.602
            \\
            5 & 4 & 3 & 10
            & 0.666 & 0.471 & 0.413
            & 0.738 & 0.556 & 0.480
            \\

            \bottomrule
        \end{tabular*}

    \caption{
        Effects of cell count, optimization, and biological source coverage
        under linear probing (LP) and fine-tuning (FT).
        The three row blocks vary these factors in sequence; all subsets
        retain three platforms.
    }
        \label{tab:data-scaling-budgets}
    \end{minipage}
\end{table*}

\begin{table*}[t!]
    \centering
    \footnotesize
    \setlength{\tabcolsep}{2.2pt}
    \renewcommand{\arraystretch}{1.00}
    \begin{tabular}{@{}ll|cccc|ccc|cccc@{}}
        \toprule
        &
        & \multicolumn{4}{|c|}{Cell}
        & \multicolumn{3}{c|}{Region}
        & \multicolumn{4}{c}{Niche} \\
        \cmidrule(lr){3-6}
        \cmidrule(lr){7-9}
        \cmidrule(lr){10-13}
        Protocol & Method
        & Brain & Liver-N & Liver-C & Lung-F
        & Liver-N & Liver-C & Lung-F
        & Brain & Liver-N & Liver-C & Lung-F \\
        \midrule

        \multirow{9}{*}{Linear probe}
        & PCA
        & 0.915 & 0.541 & 0.367 & 0.406
        & 0.445 & 0.554 & 0.383
        & 0.680 & 0.231 & 0.197 & 0.311 \\
        & CellPLM
        & 0.601 & 0.259 & 0.218 & 0.371
        & 0.218 & 0.441 & 0.310
        & 0.541 & 0.207 & 0.214 & 0.277 \\
        & NicheFormer
        & 0.895 & 0.546 & 0.430 & 0.630
        & 0.384 & 0.552 & 0.388
        & 0.687 & 0.347 & 0.354 & 0.410 \\
        & scGPT-spatial
        & 0.115 & 0.048 & 0.059 & 0.013
        & 0.189 & 0.266 & 0.098
        & 0.002 & 0.039 & 0.002 & -0.019 \\

        \cmidrule(lr){2-13}

        & CellWorld-Small
        & 0.920 & 0.622 & 0.489 & 0.714
        & 0.489 & 0.607 & 0.438
        & 0.755 & 0.392 & 0.373 & 0.460 \\
        & CellWorld-Base
        & 0.948 & 0.675 & 0.521 & 0.733
        & 0.517 & 0.634 & 0.467
        & 0.799 & 0.450 & 0.404 & 0.470 \\
        & CellWorld-Large (5\%)
        & 0.948 & 0.672 & 0.525 & 0.733
        & 0.558 & 0.668 & 0.470
        & 0.825 & 0.478 & 0.425 & 0.477 \\
        & CellWorld-Large
        & 0.952 & 0.676 & 0.526 & \textbf{0.739}
        & \textbf{0.567} & 0.679 & 0.470
        & \textbf{0.830} & 0.485 & 0.438 & 0.480 \\
        & CellWorld-Huge
        & \textbf{0.955} & \textbf{0.686} & \textbf{0.527} & 0.732
        & 0.560 & \textbf{0.693} & \textbf{0.473}
        & 0.826 & \textbf{0.504} & \textbf{0.449} & \textbf{0.481} \\

        \midrule

        \multirow{8}{*}{Fine-tuning}
        & CellPLM
        & 0.920 & 0.509 & 0.408 & 0.708
        & 0.365 & 0.557 & 0.336
        & 0.807
        & 0.458
        & 0.380
        & 0.373 \\
        & NicheFormer
        & \textbf{0.958} & \textbf{0.689} & \textbf{0.599} & \textbf{0.830}
        & 0.499 & 0.601 & 0.456
        & 0.526 & 0.122 & 0.261 & 0.313 \\
        & scGPT-spatial
        & 0.956 & 0.682 & 0.551 & 0.807
        & 0.459 & 0.595 & 0.455
        & 0.750 & 0.418 & 0.371 & 0.435 \\

        \cmidrule(lr){2-13}

        & CellWorld-Small
        & 0.951 & 0.665 & 0.505 & 0.755
        & 0.623 & 0.729 & 0.487
        & 0.825 & 0.515 & 0.446 & 0.470 \\
        & CellWorld-Base
        & 0.954 & 0.675 & 0.532 & 0.788
        & 0.622 & 0.715 & 0.498
        & 0.852 & 0.539 & 0.490 & 0.502 \\
        & CellWorld-Large (5\%)
        & 0.952 & 0.673 & 0.538 & 0.796
        & 0.643 & 0.747 & 0.502
        & 0.875 & 0.546 & 0.503 & 0.516 \\
        & CellWorld-Large
        & 0.955 & 0.677 & 0.544 & 0.792
        & 0.634 & 0.748 & \textbf{0.507}
        & \textbf{0.877} & 0.554 & 0.510 & \textbf{0.527} \\
        & CellWorld-Huge
        & 0.953 & \textbf{0.689} & 0.549 & 0.802
        & \textbf{0.670} & \textbf{0.778} & 0.505
        & 0.874 & \textbf{0.575} & \textbf{0.525} & 0.517 \\

        \bottomrule
    \end{tabular}

    \caption{
    Comparison with baselines across held-out task--dataset pairs.
    Brain, Liver-N, Liver-C, and Lung-F denote MERFISH human brain, CosMx normal liver, CosMx liver cancer, and Xenium lung fibrosis, respectively.
    CellWorld-Large (5\%) uses a broadly sampled 5\% corpus subset and the full 10k-step schedule.
    Bold marks the best result per protocol and column.
}
    \label{tab:baseline-comparison}
\end{table*}

\subsection{Comparison with Existing Methods}
We compare four full-data CellWorld scales and the 5\% CellWorld-Large variant with broad biological source coverage against three popular ST foundation models under both linear probing and fine-tuning: Nicheformer~\cite{nicheformer}, scGPT-spatial~\cite{scgpt-spatial}, and CellPLM~\cite{cellplm}.
For linear probing, we additionally include PCA, a competitive non-neural baseline that ranks among the top two external methods in 10 of 11 benchmarks.
Baseline-specific configurations are detailed in Appendix~D.
Figure~\ref{fig:baseline_radar} summarizes the comparison, while Table~\ref{tab:baseline-comparison} reports complete results for all task--dataset pairs.

Overall, CellWorld establishes a substantial performance lead over existing ST foundation models, achieving state-of-the-art results in every linear-probe benchmark and every fine-tuned spatial benchmark.
Under linear probing, even CellWorld-Small, with only 5.74M trainable
parameters, outperforms every baseline across all 11 benchmarks.
Under fine-tuning, CellWorld-Small likewise outperforms all baselines across the seven spatial benchmarks.
Performance generally improves with model scale.
For example, CellWorld's margins over the strongest baselines reach 0.157 for niche-composition prediction on CosMx normal liver under linear probing and 0.177 for region prediction on CosMx liver cancer under fine-tuning.
NicheFormer remains strongest on fine-tuned cell annotation.

Most notably, with its encoder frozen and only a linear probe trained, CellWorld-Large pretrained on just 5\% of the corpus with broad biological source coverage outperforms every fully fine-tuned baseline across all seven spatial task--dataset pairs.
On CosMx liver cancer, the frozen 5\% model leads the strongest fine-tuned baselines by 0.067 in Region F1 and 0.045 in Niche PCC.
Full-corpus pretraining further improves six of the seven spatial benchmarks, widening these margins to 0.078 and 0.058, respectively.

\section{Conclusion}
CellWorld bridges the gap between latent predictive learning and spatial transcriptomics foundation models.
Across held-out datasets, even CellWorld-Small, with 5.74M trainable parameters,
achieves state-of-the-art performance on every linear-probe benchmark and every
fine-tuned spatial benchmark, while a frozen CellWorld-Large pretrained on
only 5\% of the corpus surpasses all fully fine-tuned baselines across all seven spatial benchmarks.
Controlled scaling shows that spatial transfer improves with model capacity and depends more on sufficient optimization and broad biological source diversity than on cell count.
We hope that future large-scale perturbational and temporal ST datasets will enable CellWorld to predict how cellular states and tissue organization evolve over time and under interventions, moving toward a world model of tissue dynamics.

\bibliography{aaai2027}


\end{document}